\documentclass{article}
\usepackage{amssymb}
\usepackage{amsmath}
\usepackage{graphicx}
\usepackage{algorithm}
\usepackage{enumitem}
\usepackage{algpseudocode}
\usepackage{wrapfig}
\usepackage{caption}
\usepackage{booktabs}
\usepackage{adjustbox}
\usepackage{multirow}
\usepackage{subcaption}
\usepackage[table,xcdraw]{xcolor}

\definecolor{algo_highlight}{RGB}{38, 139, 210}
\definecolor{algo_comment}{gray}{0.5}
\definecolor{table_highlight}{rgb}{0.8, 0.9, 1}

\usepackage[preprint]{corl_2025} %

\usepackage{cleveref}
\newcommand{\refalg}[1]{\hyperref[#1]{Alg.~\ref{#1}}}   %
\newcommand{\refapp}[1]{\hyperref[#1]{App.~\ref{#1}}}   %
\newcommand{\refeq}[1]{\hyperref[#1]{Eq.~(\ref{#1})}}   %
\newcommand{\reftab}[1]{\hyperref[#1]{Tab.~\ref{#1}}}   %

\title{Robust Online Residual Refinement \\ via Koopman-Guided Dynamics Modeling}

\author{
  Zhefei Gong$^{1}$,\ Shangke Lyu$^{2\dagger}$, \ Pengxiang Ding$^{13}$,\ Wei Xiao$^{1}$, \\ 
  \textbf{Donglin Wang$^{1\dagger}$} \\
  $^1$Westlake University \ $^2$Nanjing University \ $^3$Zhejiang University\\
}

\begin{document}
\maketitle

\def\thefootnote{$\dagger$}
\footnotetext{Corresponding author. 
$^\S$Project page: \url{https://zhefeigong.github.io/korr-robot/}.
}
\def\thefootnote{\arabic{footnote}}

\begin{abstract}
    Imitation learning (IL) enables efficient skill acquisition from demonstrations but often struggles with long-horizon tasks and high-precision control due to compounding errors. 
    Residual policy learning offers a promising, model-agnostic solution by refining a base policy through closed-loop corrections. 
    However, existing approaches primarily focus on local corrections to the base policy, lacking a global understanding of state evolution, which limits robustness and generalization to unseen scenarios.
    To address this, we propose incorporating global dynamics modeling to guide residual policy updates. Specifically, we leverage Koopman operator theory to impose linear time-invariant structure in a learned latent space, enabling reliable state transitions and improved extrapolation for long-horizon prediction and unseen environments.
    We introduce \textbf{KORR} (\textbf{K}oopman-guided \textbf{O}nline \textbf{R}esidual \textbf{R}efinement), a simple yet effective framework that conditions residual corrections on Koopman-predicted latent states, enabling globally informed and stable action refinement.
    We evaluate KORR on long-horizon, fine-grained robotic furniture assembly tasks under various perturbations. Results demonstrate consistent gains in performance, robustness, and generalization over strong baselines. 
    Our findings further highlight the potential of Koopman-based modeling to bridge modern learning methods with classical control theory.

\end{abstract}

\section{Introduction}
\label{sec:introduction}

Imitation learning (IL) offers an efficient framework for acquiring task-specific policies~\citep{gong2024carp,chi2023diffusion,zhao2023learning}. Recent progress in leveraging large-scale foundation models~\citep{black2024pi0visionlanguageactionflowmodel, intelligence2025pi05visionlanguageactionmodelopenworld} has further extended IL's potential toward generalizable robotic behaviors. 
However, pure IL methods remain brittle in long-horizon, high-precision tasks like furniture assembly—common in daily life yet inherently challenging~\citep{heo2023furniturebench}. 
In such settings, even minor deviations from expert demonstrations can accumulate, leading to catastrophic failure—a problem known as compounding error. For instance, misaligning a single peg during assembly may invalidate the entire construction. Similarly, fine-grained manipulation tasks require highly accurate control, leaving little tolerance for errors.
While techniques such as action chunking~\citep{zhao2023learning} can improve consistency and fluency, they often incur long decision latencies, compromising open-loop responsiveness and increasing sensitivity to environmental perturbations.

To address these challenges, residual policy learning~\citep{silver2018residual,bi2024imitation,yuan2024policy} has emerged as a compelling solution. 
By augmenting a pre-trained base policy with corrective actions through online reinforcement learning, it provides a model-agnostic, plug-and-play refinement mechanism applicable across diverse tasks, and has demonstrated strong empirical performance. 
However, existing residual policies are typically limited to local corrections around the base policy’s output, restricting their ability to address the core challenge of robotics—robustness and generalization—an aspect largely overlooked in residual learning, 
whereas recent IL methods scale for generalization.
Conventional strategies~\citep{johannink2019residual} sample corrections only near base actions, resulting in limited global awareness and poor extrapolation to novel or unseen situations. 
Consequently, when base actions deviate substantially due to model uncertainty, residual policies often fail to recover, regardless of the base policy’s quality.

We argue that achieving robust residual learning necessitates a principled modeling of global dynamics to support effective extrapolation. In particular, modeling dynamics in a time-invariant manner introduces an essential structural prior that promotes policy stability during online residual updates.
Koopman operator theory~\citep{koopman1931hamiltonian} provides a compelling framework by lifting complex nonlinear dynamics into a linear latent space. 
In this lifted space, inherently nonlinear and coupled dynamics are represented as finite-dimensional, decoupled, and time-invariant linear transitions, which enables more reliable and globally consistent modeling of motion dynamic~\citep{mondal2023efficient}. 
It also alleviates the exponential instabilities often encountered in nonlinear systems, facilitating more stable online training.

Building on these insights, we introduce \textbf{KORR} (\textbf{K}oopman-guided \textbf{O}nline \textbf{R}esidual \textbf{R}efinement), a novel and effective framework for residual policy learning. 
KORR first models system dynamics by lifting states into a linear latent space and propagating them with learned Koopman dynamics. 
During correction, the base policy generates an action, which KORR uses to extrapolate the next imagined state. The residual policy then conditions on this state to produce a corrective action, as shown in~\autoref{fig:framework}. 
This decoupled structure enables the residual policy to leverage global information for more stable and robust refinement. By combining Koopman modeling with residual learning, KORR bridges modern learning techniques with classical control theory, offering new opportunities for robust optimization in complex tasks.
We evaluate KORR on challenging, long-horizon and fine-grained robotic tasks under various perturbations, including external disturbances and randomized initial conditions.
Experimental results show that KORR outperforms strong baselines in terms of performance, robustness, and generalization. 
Further comparisons with standard nonlinear dynamics models demonstrate the advantages of Koopman-guided modeling. 
Comprehensive ablation studies offer additional insights into the design choices behind Koopman operator learning and future study.

In summary, our contributions are as follows: 
\begin{itemize}[leftmargin=1.25em, itemsep=0em, topsep=0pt]
    \item Highlight key limitations of existing residual policy learning, particularly its reliance on local corrections and lack of global awareness, which limits robustness and generalization.
    \item Introduce KORR, a Koopman-guided residual refinement framework, 
    which enhances robustness and generalization through the extrapolation capabilities of linear time-invariant dynamics.
    \item Conduct extensive experiments and ablation studies to evaluate the effectiveness of our dynamics modeling approach, offering insights for future research.
\end{itemize}

\section{RelatedWork}
\label{sec:relatedwork}

In this section, we contextualize our work within the development of Koopman-based dynamics modeling and residual policy learning in robotics.

\textbf{Koopman Operator in Robotics}: 
The Koopman operator~\citep{koopman1931hamiltonian} offers a linear framework to analyze nonlinear dynamics by lifting them into a high-dimensional \textit{observable} space. Efficient approximations such as Dynamic Mode Decomposition (DMD)~~\citep{schmid2010dynamic} and Extended DMD (EDMD)~~\citep{williams2015data} have enabled practical estimation from time-series data. More recently, data-driven encoding approaches have been introduced~\citep{ng2023data}. 
Koopman theory has been integrated into classical control paradigms, such as LQR~\citep{brunton2016koopman, lyu2025koopman} and MPC~\citep{abraham2017model}, and later enhanced by deep learning techniques~\citep{shi2022deep}. For example, \citet{yin2022embedding} embed an LQR structure into a learned Koopman representation.
In robotics, Koopman-based modeling enables long-horizon control in interactive settings~\citep{mondal2023efficient}, and has been applied to pixel-based tasks using reinforcement learning and contrastive learning~\citep{lyu2023task, kumawat2024robokoop}, though primarily in simplified environments. Recent work extends Koopman to dexterous manipulation in both state and pixel spaces~\citep{han2023utility, chen2024korol}, but these tasks are typically short-horizon and designed for specific setups. Koopman has also been leveraged in imitation learning to improve sample efficiency via its linear constraints~\citep{bi2024imitation}.
Overall, Koopman theory has emerged as a promising modeling paradigm, facilitating learning and control across diverse robotic domains~\citep{shi2024koopman}.

\textbf{Residual Learning in Robotics}:
Residual learning has been widely adopted in deep learning to address vanishing gradients~\citep{he2016deep} and to enable efficient fine-tuning via delta updates~\citep{hu2022lora}.
In robotics, residual reinforcement learning (Residual RL) has shown promise in refining predefined controllers, particularly in challenging industrial settings~\citep{schoettler2020deep}. Residual policies have also demonstrated improved responsiveness and performance in deformable object manipulation tasks~\citep{chi2024iterative}.
Model-agnostic residual structures are increasingly used to refine base policies, including vision-language-action agents~\citep{kim2024openvla, black2024pi0visionlanguageactionflowmodel}. \citet{jiang2024transic} train residual policies via supervised learning from human preferences for sim-to-real transfer, though this requires extensive human involvement. More recent works apply online RL to train residual policies in sparse reward settings~\citep{yuan2024policy, ankile2024imitation}, showing improved task performance. 
Despite these advances, residual policies are typically constrained to operate within a narrow correction range around the base policy. 
While this design accelerates learning and convergence, it limits exploration and confines optimization to a local neighborhood—ultimately impeding the capture of global features and thereby restricting robustness and generalization.

\section{Preliminaries}
\label{sec:preliminaries}

\textbf{Koopman Operator Theory} provides a linear representation of nonlinear dynamical systems by lifting the original state space into an infinite-dimensional Hilbert space~\citep{koopman1931hamiltonian}.
Consider a discrete-time autonomous nonlinear system: 
\begin{equation} 
\mathbf{x}_{t+1} = f(\mathbf{x}_t) 
\end{equation} 
where $\mathbf{x}_t \in \mathcal{X} \subset \mathbb{R}^d$ is the system state at time $t$, and $f: \mathbb{R}^d \rightarrow \mathbb{R}^d$ denotes a nonlinear transition function.
To enable linear analysis, the Koopman framework introduces a lift function $g: \mathcal{X} \rightarrow \mathcal{O}$ that maps states into a higher-dimensional space of \textit{observables}.
The system dynamics in the lifted space evolve linearly under the action of the Koopman operator $\mathcal{K}$: 
\begin{equation}
\mathcal{K}\circ g(\mathbf{x}_t) =  g(\mathbf{x}_{t+1}) = g \circ f(\mathbf{x}_t)
\end{equation}
where $\mathcal{K}$ is a linear operator acting on the \textit{observables}.

\textbf{Koopman for Control Systems} requires one more control input $\mathbf{u}_t$ in the nonlinear discrete controlled system formulation as $\mathbf{x}_{t+1} = f(\mathbf{x}_t, \mathbf{u}_t)$. A common approach is to lift only the state $\mathbf{x}_t$ into the higher-dimensional \textit{observable} space, while leaving the control input $\mathbf{u}_t$ unchanged. Under this setting~\citep{korda2018linear, caldarelli2025linear}, the system can be represented linearly as:
\begin{equation}
g(\mathbf{x}_t, \mathbf{u}_t) = 
\begin{bmatrix}
g(\mathbf{x}_t) \\
\mathbf{u}_t
\end{bmatrix}^{\top};\ 
\mathcal{K}\circ g(\mathbf{x}_t, \mathbf{u}_t) = g(\mathbf{x}_{t+1})
\label{equation:prelim_lift_function}
\end{equation}
Operating in an infinite-dimensional space is impractical for real-world applications. Thus, Koopman analysis typically relies on finite-dimensional approximations of the operator. Specifically, the Koopman operator $\mathcal{K}$ can be approximated by a finite matrix $\boldsymbol{K} \in \mathbb{R}^{p\times p}$ as follows:
\begin{equation}
\boldsymbol{K} = 
\begin{bmatrix}
\boldsymbol{A} & \boldsymbol{B} \\
\cdot & \cdot
\end{bmatrix}
\end{equation}
where $\boldsymbol{A} \in \mathbb{R}^{m \times m}$ represents the evolution of the lifted state, and $\boldsymbol{B} \in \mathbb{R}^{m \times n}$ models the effect of the control inputs on the state evolution. 
Here, $p = m + n$, and the lower block (denoted as $(\cdot)$) represents the evolution associated with the control input $\mathbf{u}_t$, which is not explicitly modeled in this work.
The resulting Koopman-based discrete dynamics can then be expressed as:
\begin{equation}
\boldsymbol{K} \cdot g(\mathbf{x}_t, \mathbf{u}_t) = \boldsymbol{A} \cdot g(\mathbf{x}_t) + \boldsymbol{B} \cdot \mathbf{u}_t \approx g(\mathbf{x}_{t+1})
\label{equation:prelim_koopman_function}
\end{equation}

\textbf{Koopman Operator Approximation} aims to identify the optimal Koopman operator and lift function. The choice of $g$ is typically made using heuristics (e.g., polynomial lifting or kernel methods) or a neural network learned from data. 
To avoid tedious parameter tuning, we adopt the neural network approach.
Specifically, given a dataset $\mathcal{D}= \{\mathbf{x}_k,\mathbf{u}_k\}^M_{k=0}$ with $M$ total data pairs, the Koopman operator $\boldsymbol{K}$ can be obtained by minimizing the following objective function:
\begin{equation}
\boldsymbol{K} = \text{arg} \min_{\boldsymbol{K}}\frac{1}{2}\sum^{M-1}_{k=0}||g(\mathbf{x}_{k+1})-\boldsymbol{K} \cdot g(\mathbf{x}_{k},\mathbf{u}_{k})||^2
\label{equation:prelim_koopman_loss}
\end{equation}
where the optimization can be performed via least squares or data-driven approaches, such as minimizing the MSE loss using stochastic gradient descent.

\section{Method: Koopman-Guided Online Residual Refinement}
\label{sec:method}

\begin{figure}[t]
    \centering
    \includegraphics[width=0.98\textwidth]{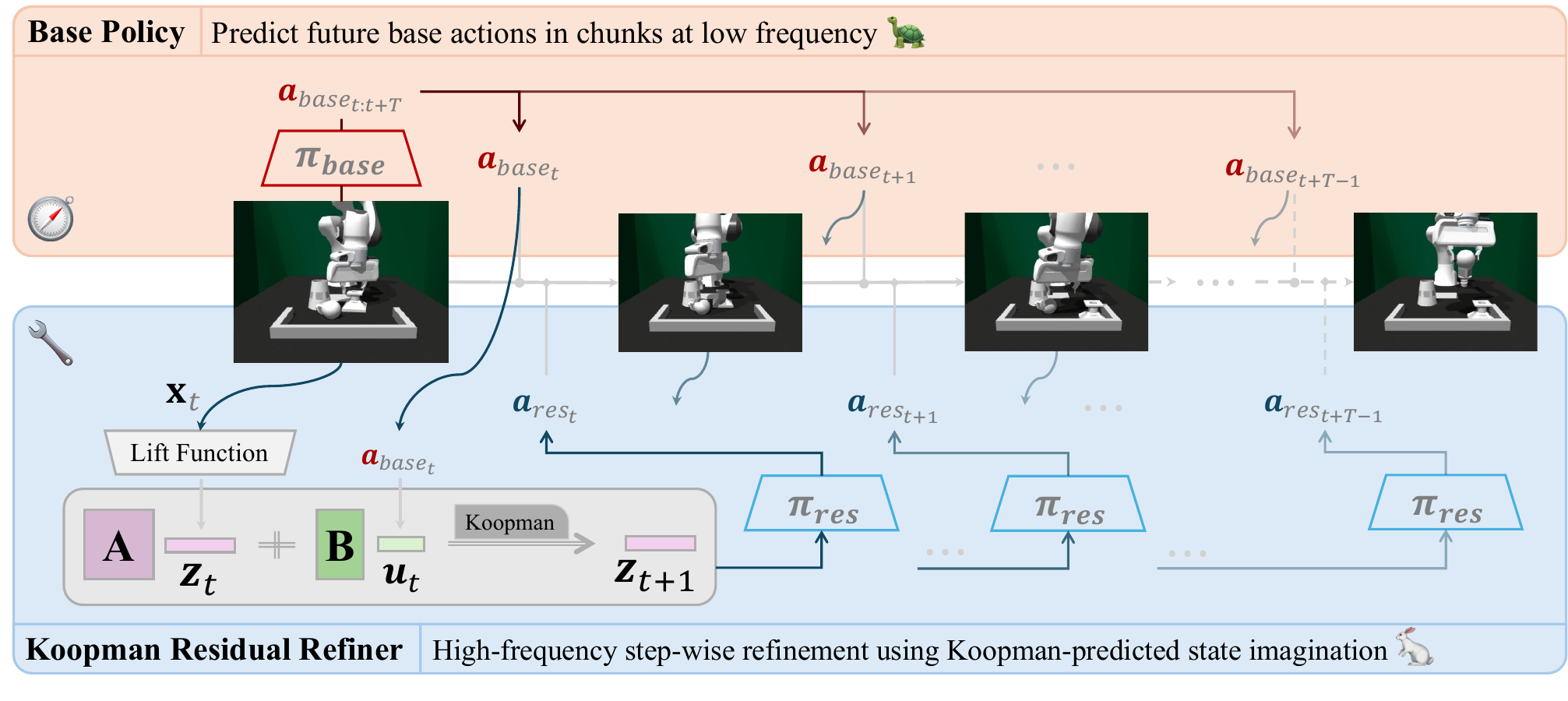}
    \caption{
    \textbf{Overview of KORR}.
    The base policy predicts a chunk of base actions at a lower frequency, while KORR refines these actions step-by-step at a higher control rate. 
    For each base action $\boldsymbol{a}_{\text{base}_t}$, KORR extrapolates the next possible state $\mathbf{z}_{t+1}$ using Koopman dynamics as an imagined future state, and then conditions the residual policy on this state to generate a corrective residual action $\boldsymbol{a}_{\text{res}_t}$. 
    The final executed action, obtained by combining the base and residual actions, forms a closed-loop refinement that enhances robustness and generalization.
    }
    \label{fig:framework}
\end{figure}

To address the limited robustness and generalization of recent conventional residual policies—and to leverage the long-horizon modeling capabilities of Koopman operator theory—we propose \textbf{KORR} (\textbf{K}oopman-guided \textbf{O}nline \textbf{R}esidual \textbf{R}efinement), a model-agnostic framework that augments any base policy with globally informed, reliable corrections. 
KORR benefits from the interpretability and extrapolation properties of Koopman-based linear dynamics, as illustrated in~\autoref{fig:framework}. Building on our problem formulation in~\autoref{sec:method_problem_form}, we develop KORR through two key components:
(1) modeling system dynamics as linear time-invariant transitions in a latent space via Koopman theory to capture structured, global transitions~(\autoref{sec:method_koopman_dynamics}); and
(2) guiding residual policy refinement using Koopman-predicted future latent states to enable stable and globally coherent corrections~(\autoref{sec:method_residual_policy}).

\subsection{Problem Formulation}
\label{sec:method_problem_form}
We formulate the task as a sequential decision-making problem: given a base policy that may suffer from imprecision or perturbations, the objective is to learn corrective residual actions to enhance execution.
We assume the base policy uses action chunking, a strategy shown to improve performance in recent work~\citep{gong2024carp, chi2023diffusion, zhao2023learning}. Notably, KORR makes no structural assumptions about the base policy, allowing it to operate with any offline-pretrained model.
At each timestep $t$, given the environment state $\mathbf{x}_t$ (or observation $\mathbf{o}_t$) and the base action $\boldsymbol{a}_{\text{base}_t}$, KORR predicts a residual correction $\boldsymbol{a}_{\text{res}_t}$.
Training is conducted via online reinforcement learning under sparse rewards, highlighting practical applicability~\citep{yuan2024policy, ankile2024imitation}.
The action space includes end-effector poses in $SE(3)$ and a binary gripper state, while the state space comprises the robot configuration (pose and velocity) and object poses.

\subsection{Koopman-Guided Dynamics Modeling}
\label{sec:method_koopman_dynamics}
KORR adopts a simple yet effective strategy by modeling forward dynamics as linear time-invariant evolution in a lifted latent space, where structured embeddings capture complex nonlinear behaviors. This design allows the residual policy to more accurately model state transitions, system structure, and residual corrections, leading to more effective refinement. Specifically, we leverage Deep Koopman embeddings to lift system states into a latent space, while assuming that a linear control term (without lifting the input) suffices for effective modeling~\citep{bruder2019modeling, brunton2016koopman}. We construct a neural network $g_{\theta}$ that maps the nonlinear state $\mathbf{x}_t \in \mathcal{X}$ into a linear Koopman \textit{observable} space $\mathbf{z}_t \in \mathcal{O}$, as shown in~\refeq{equation:prelim_lift_function}. (Alternatively, polynomial lifting up to a finite degree can also be employed.)

The Koopman operator, represented by a matrix $\mathbf{K}$, captures the linear evolution of system dynamics in the lifted latent space. Following~\refeq{equation:prelim_koopman_function}, we decompose $\mathbf{K}$ into two matrices, $\mathbf{A}$ and $\mathbf{B}$, corresponding to the contributions from the current state $\mathbf{z}_t$ and the control input $\boldsymbol{a}_t$, respectively.
\begin{equation}
\mathbf{A} \cdot \mathbf{z}_t + \mathbf{B} \cdot \boldsymbol{a}_t = \mathbf{z}_{t+1}
\label{equation:method_koopman_function}
\end{equation}
To learn the Koopman operators $\mathbf{A}$, $\mathbf{B}$, and the lift function parameters $\theta$, we optimize them via gradient backpropagation to minimize the model prediction loss $\mathcal{L}$ (see~\refeq{equation:prelim_koopman_loss}). Specifically, given a dataset $\mathcal{D} = {(\mathbf{x}_0, \boldsymbol{a}_0), \dots, (\mathbf{x}_M, \boldsymbol{a}_M)}$ of state-action trajectories, 
where $\boldsymbol{a}$ (equivalently $\boldsymbol{a}_\text{exe}$) denotes the action executed in the environment, 
the objective is to minimize the MSE loss as follows:
\begin{equation}
\mathcal{L}_{\text{kpm}} = \mathbb{E}_{t\sim\mathcal{D}}||\mathbf{z}_{t+1}-(\mathbf{A} \cdot \mathbf{z}_t+\mathbf{B} \cdot \boldsymbol{a}_t)||^2 
\ ; \ 
\mathbf{z}_{t+1} = g_{\theta}(\mathbf{x}_{t+1})
\label{equation:method_koopman_loss}
\end{equation}
Note that $\mathbf{x}_{t+1}$ represents the next state following $\mathbf{x}_{t}$ in the same trajectory. This dynamical system is time-invariant, which allows it to naturally capture manipulation skills that are more robust to intermittent perturbations compared to systems that explicitly depend on time~\citep{ravichandar2020recent}.

\subsection{Residual Policy through Koopman Imagination}
\label{sec:method_residual_policy}
Given the state $\mathbf{x}_t$ or observation $\mathbf{o}_t$ from the environment, the base policy $\pi_{\text{base}}$ generates a base action $\boldsymbol{a}_{\text{base}_t}$. Unlike traditional residual policies that generate the residual action directly based on the state and base action, we integrate an additional layer of imagination through Koopman dynamics to enhance both robustness and interpretability. Using Koopman dynamics, we first project the next imagined state $\mathbf{z}^{\text{base}}_{t+1}$ based on executing $\boldsymbol{a}_{\text{base}_t}$. Then, the residual policy $\pi_{\text{res}}$ conditions on this imagined state to generate the residual action $\boldsymbol{a}_{\text{res}_t}$(see~\refeq{equation:method_koopman_imaginary}). 
Finally, the executable action $\boldsymbol{a}_{\text{exe}_t}$ is computed by summing the base action $\boldsymbol{a}_{\text{base}_t}$ and the residual action $\boldsymbol{a}_{\text{res}_t}$.
\begin{equation}
\mathbf{A} \cdot g_{\theta}(\mathbf{x}_t) + \mathbf{B} \cdot \boldsymbol{a}_{\text{base}_t} = \mathbf{z}^{\text{base}}_{t+1}
\ ;\ 
\boldsymbol{a}_{\text{res}_t} = \pi_{\text{res}}(\mathbf{z}^{\text{base}}_{t+1})
\label{equation:method_koopman_imaginary}
\end{equation}
We train the efficient residual Multi-Layer Perceptron (MLP) policy $\pi_{\text{res}}$ with the base policy frozen, using online RL to fine-tune and address issues of imprecision, slow reactivity, and limited robustness and generalization. We apply PPO~\citep{schulman2017proximal} (see~\refapp{appendix:ppo_alg} for details). With these straightforward updates (highlighted in \textcolor{algo_highlight}{light-blue} in~\refalg{algo:koopman_residual_policy}), we leverage the robustness of Koopman dynamics' extrapolation for a holistic system view. Additionally, by conditioning on the predicted next state rather than the base action and current state, we decouple the residual policy’s understanding, mirroring human decision-making, which often relies on anticipated outcomes over direct sensory inputs~\citep{summerfield2014expectation}.

\begin{algorithm}
\caption{KORR pseudo-code with PPO}
\label{algo:koopman_residual_policy}
\begin{algorithmic}[1]  %
\State Initialize base policy $\pi_{\text{base}}$ (pretrained and frozen here), residual policy $\pi_{\text{res}}$ with parameters $\phi$.
\State \textcolor{algo_highlight}{Initialize Koopman Operator $\mathbf{A}$ and $\mathbf{B}$, and lift function $g(\cdot)$ with parameters $\theta$. }
\For{iteration = 1 to N}
    \For{each environment rollout step}
        \State Observe current state $\mathbf{x}_t$
        \State Compute base action: $\boldsymbol{a}_{\text{base}_t} = \pi_{\text{base}}(\mathbf{x}_t)$
        \State \textcolor{algo_highlight}{Koopman extrapolate: $\mathbf{A}\cdot g_{\theta}(\mathbf{x}_t)+\mathbf{B} \cdot \boldsymbol{a}_{\text{base}_t}=\mathbf{z}^{\text{base}}_{t+1}$ }
        \textcolor{algo_comment}{\Comment{See description in~\refeq{equation:method_koopman_imaginary}}}
        \State Compute residual action based the imaginary next state: $\boldsymbol{a}_{\text{res}_t} = \pi_{\text{res}}(\textcolor{algo_highlight}{\mathbf{z}^{\text{base}}_{t+1}})$
        \State Execute $\boldsymbol{a}_{\text{exe}_t}=\boldsymbol{a}_{\text{base}_t}+\boldsymbol{a}_{\text{res}_t}$, observe reward $r$, next state $\mathbf{x}_{t+1}$
        \State Store $(\mathbf{x}_t, \boldsymbol{a}_{\text{exe}_t}, r, \mathbf{x}_{t+1})$ into buffer $\mathcal{D}$
    \EndFor
    \State Compute advantage estimates $A$ using GAE \textcolor{algo_comment}{\Comment{See~\refapp{appendix:ppo_alg}}}
    \For{update step}
        \State \textcolor{algo_highlight}{Compute Koopman loss $\mathcal{L}_{\text{kpm}}$ to update $\theta$, $\mathbf{A}$, and $\mathbf{B}$}
        \textcolor{algo_comment}{\Comment{MSE loss depicted in~\refeq{equation:method_koopman_loss}}}
        \State Compute PPO loss $\mathcal{L}_{\text{ppo}}$ to update all of the parameters including the $\phi$
        \textcolor{algo_comment}{\Comment{See~\refapp{appendix:ppo_alg}}}
    \EndFor
\EndFor
\end{algorithmic}
\end{algorithm}

\section{Experiment}
\label{sec:experiment}

With the above relatively simple modifications, our method, \textbf{KORR}, demonstrates stronger robustness, generalization, and higher performance compared to conventional residual policies under online reinforcement learning. In this section, we present experimental results designed to address the following research questions (\textbf{RQ}s):
\begin{itemize}[leftmargin=1.25em, itemsep=0em, topsep=0pt] 
    \item \textbf{RQ1}: Does KORR improve robustness and performance over traditional residual policies?
    \item \textbf{RQ2}: Does Koopman modeling offer advantages over nonlinear dynamics for residual learning?
    \item \textbf{RQ3}: Which design choices in KORR are critical for stable performance?
\end{itemize}

\subsection{Experimental Design}
\label{sec:experiment_exp_des}
\textbf{Benchmarks and Baselines}: We evaluate KORR on the challenging 6-Degree-of-Freedom (DoF) long-horizon \textit{Furniture Assembly} benchmark~\citep{heo2023furniturebench}, built with IsaacGym~\citep{makoviychuk2021isaac}. This benchmark has been widely studied in recent works~\citep{ankile2024juicer, ankile2024imitation, jiang2024transic, lin2024generalize} for its real-world complexity and fine-grained manipulation skills.
For base policies, we adopt the official implementations of (1) \textit{Diffusion Policy (DP)}~\citep{chi2023diffusion}, which learns via denoising diffusion probabilistic models~\citep{ho2020denoising}, and (2) \textit{CARP}~\citep{gong2024carp}, which employs coarse-to-fine autoregressive prediction. 
To evaluate residual policy designs, we compare KORR against: 
(1) \textit{ResiP}~\citep{ankile2024imitation}, a basic MLP-based residual policy; and 
(2) \textit{ResiP + Non-Linear Dynamics}, a variant where residuals are predicted via learned nonlinear dynamics (MLP). 
Residual policies are independently trained for each base policy–task pair (see~\refapp{appendix:implementation_details} for additional details).

\begin{wrapfigure}{r}{0.64\textwidth}
  \vspace{-6mm}
  \begin{center}
    \includegraphics[width=0.99\linewidth]{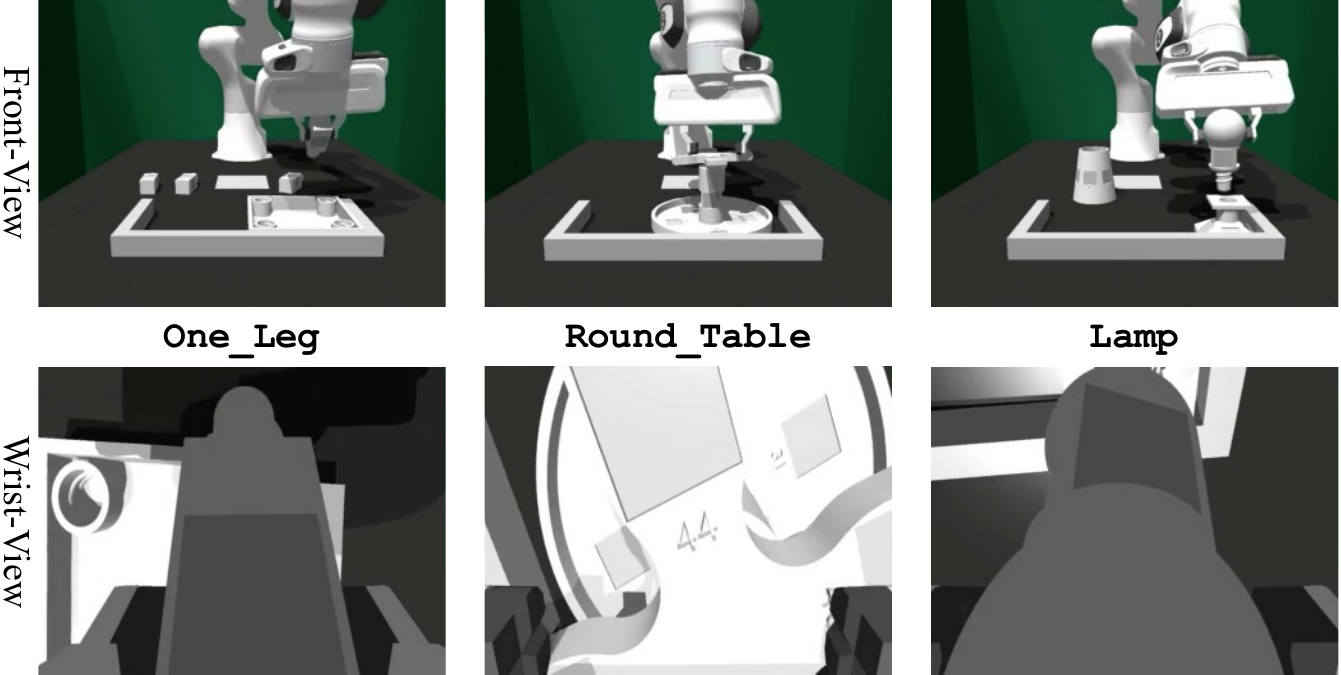}
  \end{center}
  \vspace{-0.1in}
  \caption{\textbf{Task visualizations.}
  Snapshots of the IKEA Furniture Assembly tasks used for evaluation throughout our experiments.}
  \label{fig:task_vis}
  \vspace{-4mm}
\end{wrapfigure}
\textbf{Tasks and Metrics}:
We evaluate on three challenging tasks from the IKEA Furniture Assembly benchmark~\citep{heo2023furniturebench}: \texttt{One\_Leg}, \texttt{Round\_Table}, and \texttt{Lamp} (visualized in~\autoref{fig:task_vis}). Each task requires executing long-horizon dexterous skills such as picking, placing, inserting, screwing, and flipping, over episodes lasting up to 1000 steps (\texttt{One\_Leg} is capped at 700 steps). 
To systematically assess robustness and generalization, we introduce (1) \textit{initial randomness} in object placements, categorized into \texttt{Low}, \texttt{Med}, and \texttt{High} levels, and (2) \textit{external disturbances}, where objects are perturbed after each action (\texttt{w/o} vs. \texttt{w/} disturbance). These settings simulate real-world uncertainties (see~\refapp{appendix:benchmark_task_settings} for details). Policies are evaluated over 1024 episodes, and success rates are reported based on the best-performing checkpoint.

\subsection{General Robustness and Performance Study}
\label{sec:experiment_general_study}
To thoroughly evaluate residual policy performance, we introduce two levels of initialization randomness—\texttt{Low} and \texttt{Med}—with an additional \texttt{High} level used exclusively for the \texttt{One\_Leg} task. 
Each level is associated with a disturbance scale, where scene components are randomly perturbed after each interaction with the environment to simulate the robustness required in real-world tasks.
All residual policies are fine-tuned via reinforcement learning on a fixed, disturbance-free base policy, with disturbances applied only during the evaluation phase to fully test the model’s robustness. 
Here, we denote evaluation without disturbances as \texttt{w/o}, and with level-specific disturbances as \texttt{w/}.

As shown in~\autoref{tab:general_robustness_study}, KORR not only consistently improves success rates under normal conditions but also exhibits remarkable robustness across all tasks and randomness levels when disturbances are applied. 
The impact of Koopman dynamics plays a key role in this improvement. By modeling the global dynamics, KORR enables the policy to generalize effectively, even under unseen conditions. This extrapolation capability is especially advantageous when disturbances create novel situations. These results strongly support \textbf{RQ1}, demonstrating that KORR achieves substantial performance and robustness improvements with minimal changes to the implementation, as illustrated in~\refalg{algo:koopman_residual_policy}.

\begin{table*}[t]
\small
\centering
\setlength{\tabcolsep}{3pt}
\begin{adjustbox}{center}
\begin{tabular}{lcccccccccccccc}
\toprule
\multirow{2}{*}{\textbf{Methods}} 
& \multicolumn{6}{c}{\texttt{One\_Leg}} 
& \multicolumn{4}{c}{\texttt{Round\_Table}} 
&  \multicolumn{4}{c}{\texttt{Lamp}} 
\\
\cmidrule(r){2-7} 
\cmidrule(r){8-11} 
\cmidrule(r){12-15}
& \multicolumn{2}{c}{\texttt{Low}} 
& \multicolumn{2}{c}{\texttt{Med}} 
& \multicolumn{2}{c}{\texttt{High}} 
& \multicolumn{2}{c}{\texttt{Low}} 
& \multicolumn{2}{c}{\texttt{Med}} 
& \multicolumn{2}{c}{\texttt{Low}} 
& \multicolumn{2}{c}{\texttt{Med}} \\
\midrule
DP~\citep{chi2023diffusion}
& \multicolumn{2}{c}{47.97}  
& \multicolumn{2}{c}{23.93}  
& \multicolumn{2}{c}{4.39}  
& \multicolumn{2}{c}{5.76}  
& \multicolumn{2}{c}{1.56} 
& \multicolumn{2}{c}{5.57}  
& \multicolumn{2}{c}{1.46}  \\
\midrule
& \texttt{w/o} & \texttt{w/} & \texttt{w/o} & \texttt{w/} & \texttt{w/o} & \texttt{w/} & \texttt{w/o} & \texttt{w/} & \texttt{w/o} & \texttt{w/} & \texttt{w/o} & \texttt{w/} & \texttt{w/o} & \texttt{w/} \\
\midrule
ResiP~\citep{ankile2024imitation} 
& 98.14 & 85.35 & 84.99 & 46.09 & 30.03 & 3.20 & 96.19 & 80.27 & 52.73 & 22.66 & 86.58 & 60.55 & 51.66 & 29.98 \\
\rowcolor{table_highlight}
KORR (ours)  
& 98.73 & \textbf{90.38} & \textbf{87.21} & \textbf{52.31} & \textbf{44.04} & \textbf{8.30} & 96.78 & 81.35 & \textbf{67.19} & \textbf{27.25} & \textbf{89.65} & \textbf{63.48} & \textbf{74.47} & \textbf{39.16} \\
\bottomrule
\end{tabular}
\end{adjustbox}
\caption{ \textbf{Performance comparison across tasks and difficulty levels.} We report the success rate (\%) based on 1024 rollouts for each task. The residual policies (ResiP and KORR) are built upon the same base policy (DP), with the entire training process conducted without disturbances. Here, \texttt{w/o} denotes evaluation without disturbances, while \texttt{w/} refers to evaluation with the corresponding level of disturbance. 
By seamlessly incorporating simple Koopman dynamics, KORR consistently surpasses conventional residual policies in both task performance and robustness.
}
\label{tab:general_robustness_study}
\vspace{-4mm}
\end{table*}

\begin{wrapfigure}{l}{0.60\textwidth}
    \vspace{-4mm}
    \small
    \centering
    \setlength{\tabcolsep}{4pt}
    \begin{tabular}{lcccccc}
    \toprule
    \multirow{2}{*}{\textbf{Methods}} & 
    \multicolumn{2}{c}{\texttt{One\_Leg}} & 
    \multicolumn{2}{c}{\texttt{Round\_Table}} & 
    \multicolumn{2}{c}{\texttt{Lamp}} \\
    \cmidrule(r){2-3} 
    \cmidrule(r){4-5} 
    \cmidrule(r){6-7} 
    & \texttt{$\rightarrow$Med}
    & \texttt{$\downarrow \Delta$}
    & \texttt{$\rightarrow$Med}
    & \texttt{$\downarrow \Delta$}
    & \texttt{$\rightarrow$Med}
    & \texttt{$\downarrow \Delta$} \\
    \midrule
    ResiP~\citep{ankile2024imitation} & 16.41 & 83.28 & 7.32 & 92.39 & 17.28 & 80.04 \\
    \rowcolor{table_highlight}
    KORR(ours) & \textbf{22.27} & 77.44 & \textbf{10.54} & 89.11 & \textbf{18.75} & 79.09 \\
    \bottomrule
    \end{tabular}
    \captionof{table}{ 
    \textbf{Generalization performance.}
    Direct evaluation of \texttt{Low}-trained policies under \texttt{Med} initialization randomness.
    }
    \label{tab:initial_randomness_study}
    \vspace{-4mm}
\end{wrapfigure}
We further assess the generalization capabilities of KORR enabled by Koopman dynamics. In this experiment, residual policies trained under \texttt{Low} initialization randomness are directly deployed at the \texttt{Med} level, evaluating their ability to adapt to previously unseen scenarios, while the base policy remains frozen throughout. 
As shown in~\reftab{tab:initial_randomness_study}, KORR consistently achieves stronger generalization, exhibiting higher success rates at \texttt{Med} (measured by $\rightarrow$ \texttt{Med}) and smaller relative performance drops (measured by $\downarrow \Delta$, computed as $\frac{\Delta x}{x_{\text{original}}} \times 100\%$), compared to ResiP. These results further validate KORR's robust extrapolation to unseen scenarios, thus strongly supporting \textbf{RQ1}.

\begin{wrapfigure}{r}{0.78\textwidth}
  \vspace{-6mm}
  \begin{center}
    \includegraphics[width=\linewidth]{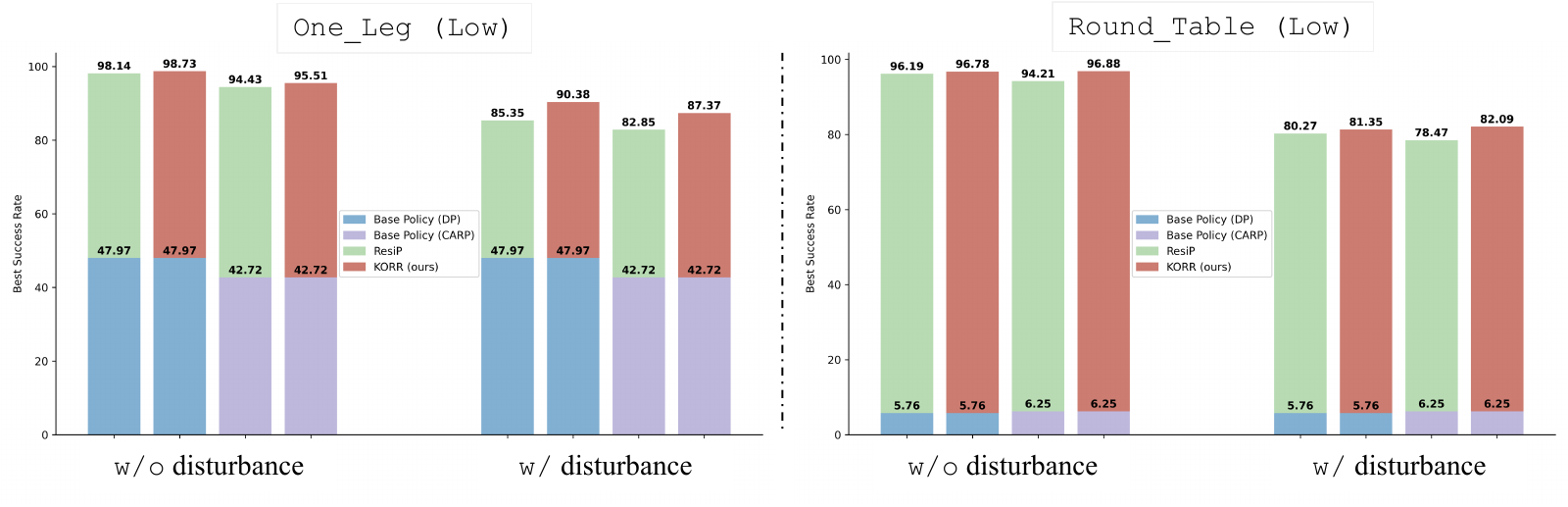}
  \end{center}
  \vspace{-0.15in}
  \caption{
  \textbf{Universality across different base policies.}
  KORR consistently enhances performance and robustness with an alternative base policy.}
  \label{fig:diff_base_policy}
  \vspace{-4mm}
\end{wrapfigure}
We further evaluate the universality of KORR by applying it to different base policies, retrained for each scenario.
In ~\autoref{fig:diff_base_policy}, KORR, when applied to an autoregressive base policy, CARP~\citep{gong2024carp}, consistently outperforms legacy residual methods, achieving significant improvements in both success rate and robustness. 
These findings provide further support for \textbf{RQ1} and demonstrate KORR's flexibility in integrating with various base policies. 
Additionally, we observe that the initial quality of the base policy influences the effectiveness of residual learning, as detailed in~\refapp{appendix:influ_base_to_residual}. 
The reported success rates of the base policies reflect the frozen performance under standard, disturbance-free conditions.

\subsection{Linear Benefit Study}
\label{sec:experiment_linear_study}

\begin{figure}[t]
    \centering
    \begin{subfigure}[t]{0.58\textwidth}
        \centering
        \includegraphics[width=\textwidth]{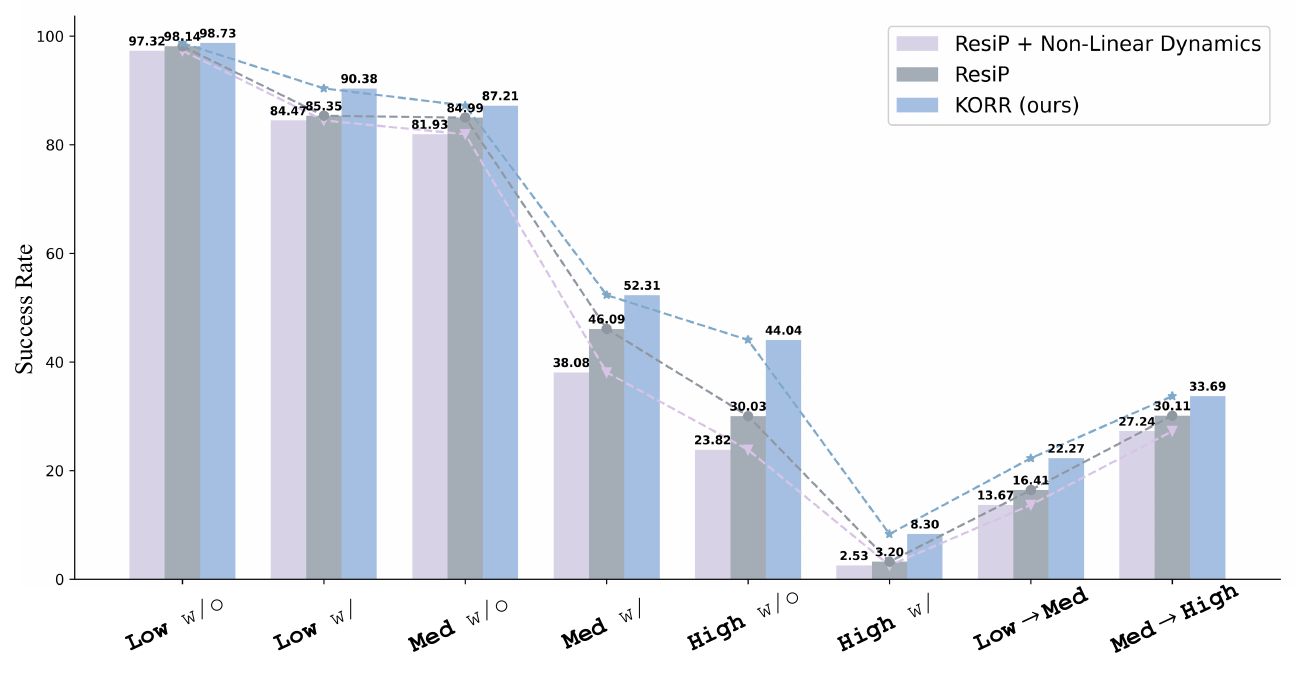}
        \caption{Success rate in \texttt{One\_Leg} task.}
        \label{fig:comp_korr_resip_w_dyn}
    \end{subfigure}
    \hfill
    \begin{subfigure}[t]{0.40\textwidth}
        \centering
        \includegraphics[width=\textwidth]{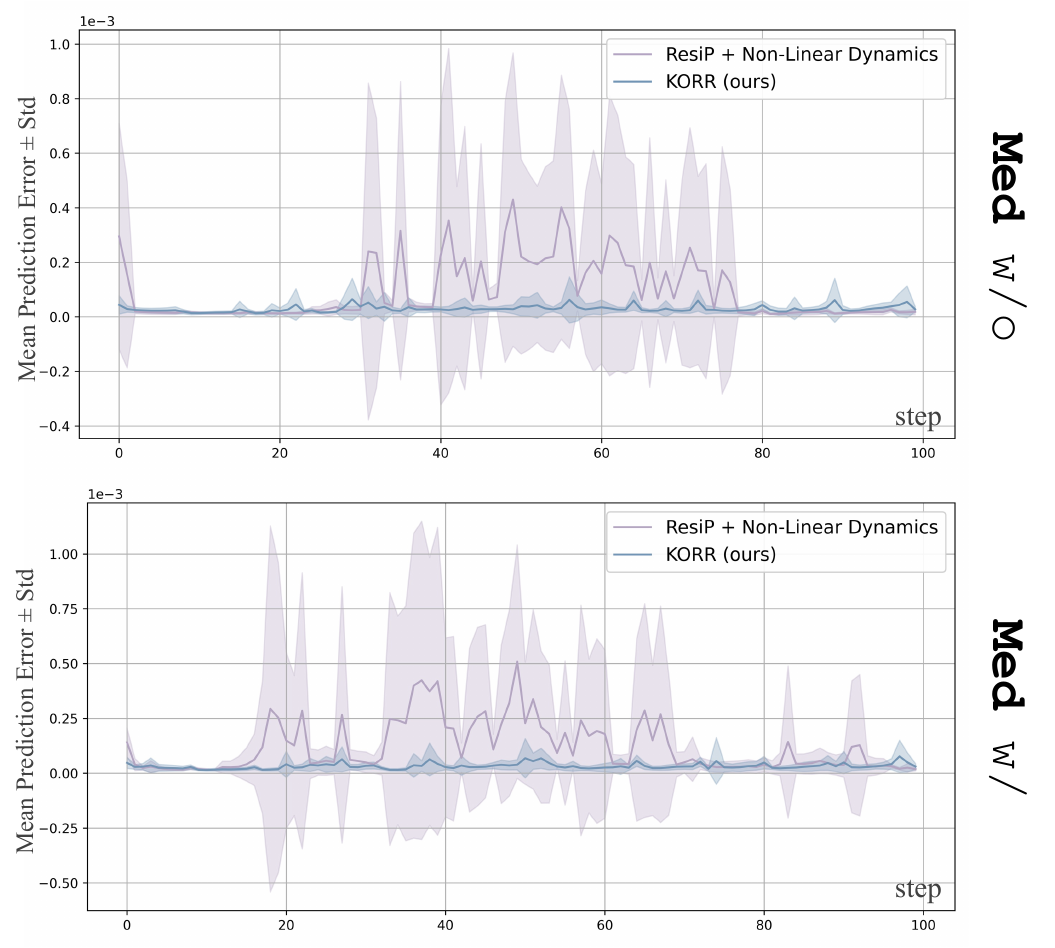}
        \caption{Dynamics prediction error.}
        \label{fig:comp_korr_resip_w_dyn_vis}
    \end{subfigure}
    \caption{
    \textbf{Comparison between non-linear and Koopman-guided dynamics.} 
    (a) shows that non-linear dynamics hurt performance, while Koopman-based dynamics consistently improve robustness and generalization. 
    (b) shows the stability advantage of linear Koopman constraints, yielding more consistent and accurate long-term extrapolations, measured by MSE (mean ± std) over 10 rollouts.
    }
    \label{fig:combined_fig}
    \vspace{-2mm}
\end{figure}

Compared to traditional residual policies~\citep{ankile2024imitation}, KORR incorporates a simple yet effective linear dynamics model during online RL fine-tuning.
In contrast, conventional designs typically favor non-linear dynamics, often parameterized by MLPs. 
To better isolate the benefit of linear structure, we introduce an additional baseline, \textit{ResiP + Non-Linear Dynamics} (details in~\refapp{appendix:resip_w_nlin_dyn}), where a non-linear transition dynamics model predicts future states conditioned on the current state and base action, and the residual policy is trained accordingly on these imagined outcomes as well.

As shown in~\autoref{fig:comp_korr_resip_w_dyn}, we evaluate performance across all levels of randomness in the \texttt{One\_Leg} task, under both \texttt{w/o} and \texttt{w/} disturbance conditions, as well as in generalization settings (\texttt{Low}/\texttt{Med} training to \texttt{Med}/\texttt{High} evaluation). The results show that KORR consistently outperforms the ResiP baseline, whereas the addition of non-linear dynamics leads to performance degradation. 
Compared to non-linear extrapolation, KORR enforces a linear time-invariant structure through Koopman modeling, encouraging the learning of globally consistent dynamics. This inductive bias improves long-horizon extrapolation, mitigates overfitting to spurious correlations, and reduces cumulative prediction errors—ultimately enhancing robustness and generalization, thus supporting \textbf{RQ2}.
In~\autoref{fig:comp_korr_resip_w_dyn_vis}, KORR achieves 
significantly lower prediction drift over time, whereas non-linear dynamics often exhibit instability and oscillation (see \refapp{appendix:linear_vs_nonlinear}).
This linear prior provides stable inductive regularization and helps avoid the divergence commonly observed in unconstrained non-linear models.

\subsection{Additional Ablation Study}
\label{sec:experiment_ablation_study}
To further investigate the design choices of KORR and address \textbf{RQ3}, we conduct a series of further studies focusing on the use of Koopman-guided dynamics.
We analyze the impact of the exploration rate during residual policy training (\refapp{appendix:explo_rate_res_learn}) and examine how the quality of the base policy influences the learning of residuals (\refapp{appendix:influ_base_to_residual}).
Regarding the Koopman dynamics itself, we perform extensive ablation studies (\refapp{appendix:add_abl_exp}), including evaluations of the lifted space dimension (\refapp{appendix:add_abl_exp_liftdim}), the importance of backpropagating the RL loss into the Koopman module (\refapp{appendix:add_abl_exp_bkp}), and the feasibility of incorporating goal-conditioning (\refapp{appendix:add_abl_exp_goal}). These comprehensive ablations offer strong guidance for the future application and refinement of the Koopman-guided framework.

\section{Conclusion}
\label{sec:conclusion}

In this work, we presented KORR, a Koopman-guided residual refinement framework that enhances robustness and generalization by modeling global dynamics through linear time-invariant structures induced by Koopman operator theory in a lifted latent space. 
By constraining residual learning within this linearized space, KORR improves stability and mitigates the sensitivity of non-linear dynamics to disturbances. 
Comprehensive experiments and ablation studies further validate the design choices and offer guidance for future extensions.
Moreover, by leveraging Koopman-based linearization, KORR naturally connects learning-based approaches with classical control techniques, such as LQR~\citep{shi2022deep} and ESO~\citep{Han2009ADRC}, which are inherently effective in linear regimes.
We believe KORR represents a promising step toward integrating control-theoretic insights into learning frameworks, paving the way for more robust and principled learning in future practical robotics applications.

\clearpage
\section{Limitation}
\label{sec:limitation}

\textbf{Limitations} remain, despite KORR’s enhanced robustness and generalization through Koopman-based linear dynamics:
First, the assumption that system dynamics can be approximated as linear in the lifted Koopman space may break down in highly non-stationary or contact-rich environments, leading to inaccurate predictions and suboptimal corrections.
Second, the Koopman model in KORR is learned purely from data, without incorporating physical or heuristic priors. This can limit its generalization in low-data regimes, where structured inductive biases may be beneficial.
Third, although KORR is designed to be model-agnostic, its performance can still be affected by the quality and coverage of the base policy, as further discussed in~\refapp{appendix:influ_base_to_residual}.
Fourth, KORR relies on online reinforcement learning under sparse rewards, which—while reflecting realistic robotic settings—results in high sample complexity, often requiring hundreds of millions of interactions. This training inefficiency currently limits applicability to vision-based or real-world scenarios. 
Finally, although KORR operates in a state-based setting with a small sim-to-real gap, yielding consistent results between simulation and the real world, the primary limitation remains accurate 6D object state estimation. When predictions are precise, real-world performance should closely match simulation. Nevertheless, our experiments are mainly conducted in simulation, and transferring KORR to real-world systems may still face challenges such as sim-to-real discrepancies, latency, and more complex disturbance patterns.
\textbf{Future Work} may explore:
(i) incorporating physics-informed or heuristic priors to improve the fidelity of Koopman approximations in more complex scenarios, such as dynamics-level disturbances;
(ii) developing sample-efficient residual learning techniques, such as selective correction mechanisms based on base policy confidence; 
(iii) leveraging the linear structure of Koopman representations to incorporate classical control methods (e.g., robust control), promoting stable residual closed-loop correction and improving sim-to-real transfer.

\clearpage
\bibliography{sec/y_references}  %

\clearpage
\appendix
\textbf{\Large{Appendix}}
\vspace{5pt}

\section{Environment Implementation Details}
\label{appendix:implementation_details}
We adopt the official implementations of the base policies and list key hyperparameters in~\reftab{tab:para_dp} and~\reftab{tab:para_carp}. 
Across all policies, we maintain consistent temporal settings: an observation (or state) horizon of 1 and an action executing horizon of 8. 
This short-horizon observation configuration is easy to align with real-world deployment constraints, minimizing latency and ensuring responsiveness.
Regarding the action space, we follow prior works~\citep{chi2023diffusion, gong2024carp} and use \textit{absolute} end-effector poses as actions. 
For orientation representation, we employ the 6D continuous rotation representation~\citep{zhou2019continuity}, which is known to facilitate stable training, compared to Euler or quaternion representations.

\begin{table*}[ht]
\centering
\small
\begin{minipage}[b]{0.48\textwidth}
    \centering
    \begin{adjustbox}{center}
    \begin{tabular}{lc}
        \toprule
        Parameter & Value\\
        \midrule
        net\_name  & unet \\
        down\_dims & [256, 512, 1024] \\
        diffusion\_step\_embed\_dim & 256\\
        kernel\_size & 5 \\
        num\_groups & 8 \\
        num\_diffusion\_iters & 100 \\
        num\_inference\_steps & 16 \\
        clip\_sample & true \\
        prediction\_type & epsilon \\
        beta\_schedule & squaredcos\_cap\_v2 \\
        pred\_horizon & 32 \\
        obs\_horizon & 1 \\
        action\_horizon & 8 \\
        batch\_size & 256 \\
        \bottomrule
    \end{tabular}
    \end{adjustbox}
    \caption{
    \textbf{Hyperparameters of DP~\citep{chi2023diffusion}.}
    }
    \label{tab:para_dp}
\end{minipage}
\hfill
\begin{minipage}[b]{0.48\textwidth}
    \centering
    \begin{adjustbox}{center}
    \begin{tabular}{lc}
        \toprule
        Parameter & Value\\
        \midrule
        patch\_nums  & [1, 3, 6, 8] or [1, 2, 3, 4] \\
        vocab\_size & 512 \\
        vocab\_ch & 8 \\
        vch & 2 \\
        vqnorm & true \\
        act\_dim\_names & [x,y,z,r1,r2,r3,r4,r5,r6,gripper] \\
        act\_dim & 10 \\
        tdepth & 16 \\
        tembed & 64 \\
        opt & adamw \\
        pred\_horizon & 32 or 16\\
        obs\_horizon & 1\\
        action\_horizon & 8\\
        batch\_size & 256 \\
        \bottomrule
    \end{tabular}
    \end{adjustbox}
    \caption{
    \textbf{Hyperparameters of CARP~\citep{gong2024carp}.}
    }
    \label{tab:para_carp}
\end{minipage}
\end{table*}

For online reinforcement learning, we list the key hyperparameters used in~\reftab{tab:para_online}. Apart from the additional observable lifting dimension \textit{obs\_lift\_dim} in KORR, all configurations follow the setup of PPO-based ResiP training~\citep{ankile2024imitation}, with only minimal modifications.

\begin{table}[ht]
    \centering
    \small
    \begin{adjustbox}{center}
    \begin{tabular}{lclc}
        \toprule
        Parameter & Value & Parameter & Value \\
        \midrule
        obs\_lift\_dim & 256 & init\_logstd & -1.0 \\
        learn\_std & false & critic\_last\_layer\_activation & None \\
        action\_head\_std & 0.0 & critic\_last\_layer\_std & 0.25 \\
        action\_scale & 0.1 & critic\_last\_layer\_bias\_const & 0.25 \\
        actor\_num\_layers & 2 & critic\_num\_layers & 2 \\
        actor\_hidden\_size & 256 & critic\_hidden\_size & 256  \\
        actor\_activation & ReLU & critic\_activation & ReLU  \\
        actor\_bias\_on\_last\_layer & false & critic\_bias\_on\_last\_layer & true \\
        \bottomrule
    \end{tabular}
    \end{adjustbox}
    \vspace{0.1in}
    \caption{
    \textbf{Key hyperparameters for online fine-tuning of residual policies.}
    }
    \label{tab:para_online}
\end{table}

\section{Benchmark Task Settings}
\label{appendix:benchmark_task_settings}
We structure the initialization randomness across several levels for each task. For the tasks \texttt{Round\_Table} and \texttt{Lamp}, we use two levels of randomness: \texttt{Low} and \texttt{Med}. For the \texttt{One\_Leg} task, we introduce three levels: \texttt{Low}, \texttt{Med}, and \texttt{High}, with \texttt{High} representing a greater degree of randomness during initialization. All tasks are executed in the Isaac Gym environment~\citep{makoviychuk2021isaac}, with the underlying action execution frequency set to 10 Hz.

\begin{figure}[ht]
    \centering
    \includegraphics[width=1.0\textwidth]{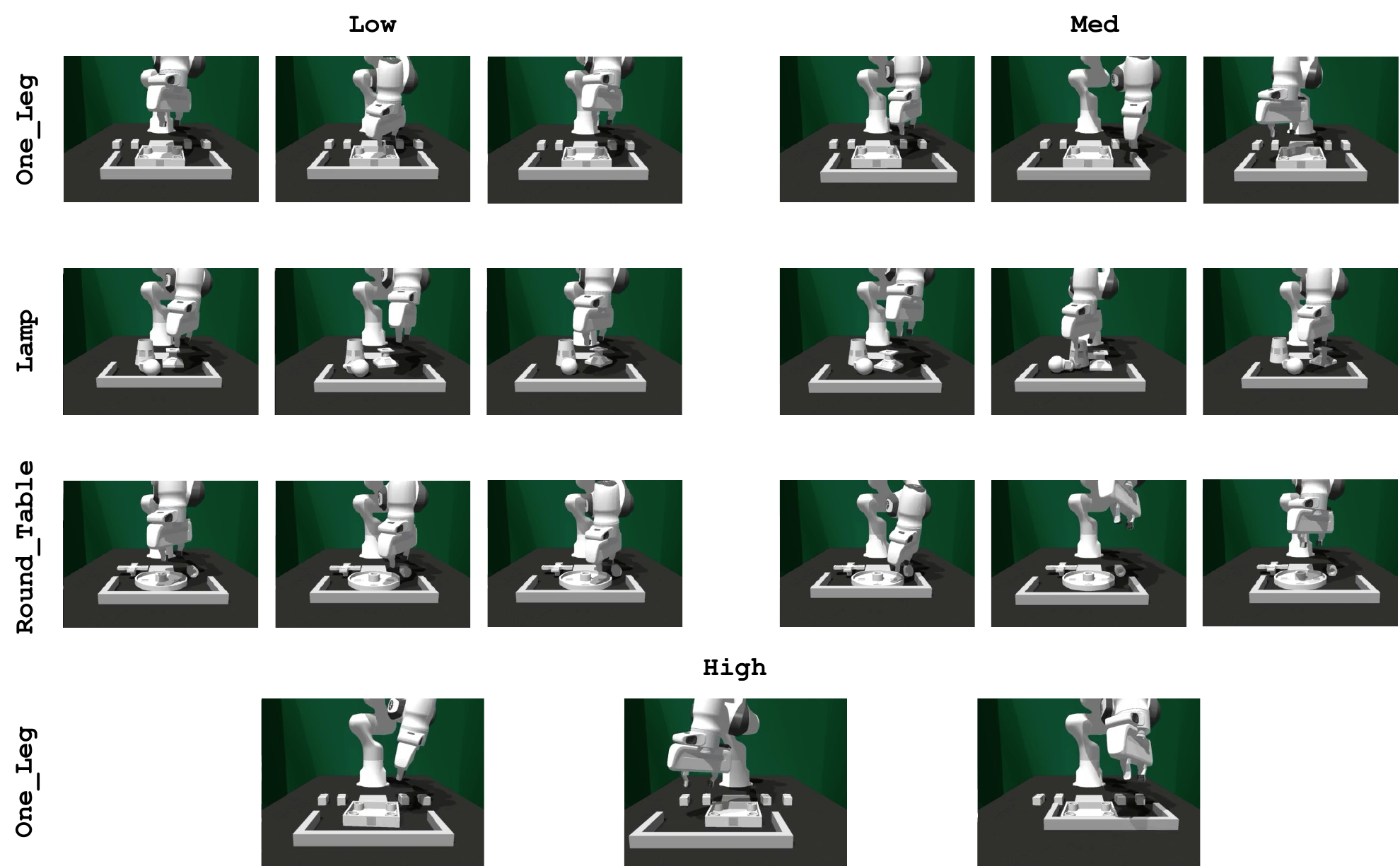}
    \caption{
    \textbf{Visualization of initialization randomness for each task.}
    }
    \label{fig:ini_rand_level}
\end{figure}

For each level of initialization randomness, we first place the object in a designated area suitable for task execution, introducing slight random perturbations to the obstacle and the Franka robot arm. 
We then introduce controlled disturbances to the position and orientation of all objects within predefined ranges, following the methodology outlined in~\citet{ankile2024imitation}, to induce the desired level of randomness.
The disturbance scales and initialization parameters are detailed in~\reftab{tab:para_disturb}. The visualization of the initialization randomness for each task and level is shown in~\autoref{fig:ini_rand_level}. 
These levels of randomness are designed to reflect practical real-world generalization challenges, enabling a comprehensive evaluation of the model’s performance under diverse conditions.

\begin{table}[ht]
    \centering
    \small
    \begin{adjustbox}{center}
    \begin{tabular}{lcccc}
        \toprule
        Hyperparameters & \texttt{Low} & \texttt{Med} & \texttt{High} & Note\\
        \midrule
        \text{max\_force\_magnitude}  & 0.2  & 0.5 & 0.75 &
        translation disturbance \\
        \text{max\_torque\_magnitude} &  0.007 & 0.01 & 0.015 &
        rotation disturbance \\
        \text{max\_obstacle\_offset} & 0.02 & 0.04 & 0.06 &
        obstacle reset initialization \\
        \text{franka\_joint\_rand\_lim\_deg} & $\text{radians}(5^\circ)$ & $\text{radians}(10^\circ)$ & $\text{radians}(13^\circ)$ & 
        joint angle deviation initialization\\
        \bottomrule
    \end{tabular}
    \end{adjustbox}
    \vspace{0.1in}
    \caption{
    \textbf{Disturbance and initialization parameters.} 
    Three levels of disturbance and initialization are defined as \texttt{Low}, \texttt{Med}, and \texttt{High}. The note provides a basic explanation.
    } 
    \label{tab:para_disturb}
\end{table}

To further assess the robustness of the residual policy, we simulate disturbances during task execution, reflecting real-world deployment scenarios. Following the disturbance scales in~\reftab{tab:para_disturb}, we introduce additional position and orientation perturbations to the objects after each interaction with the environment. The severity of these disturbances increases in line with the initial randomness levels to fully test the policy’s adaptability to dynamic, unpredictable conditions.

\section{Residual Policy with Non-Linear Dynamics}
\label{appendix:resip_w_nlin_dyn}
\begin{figure}[ht]
    \centering
    \includegraphics[width=0.8\textwidth]{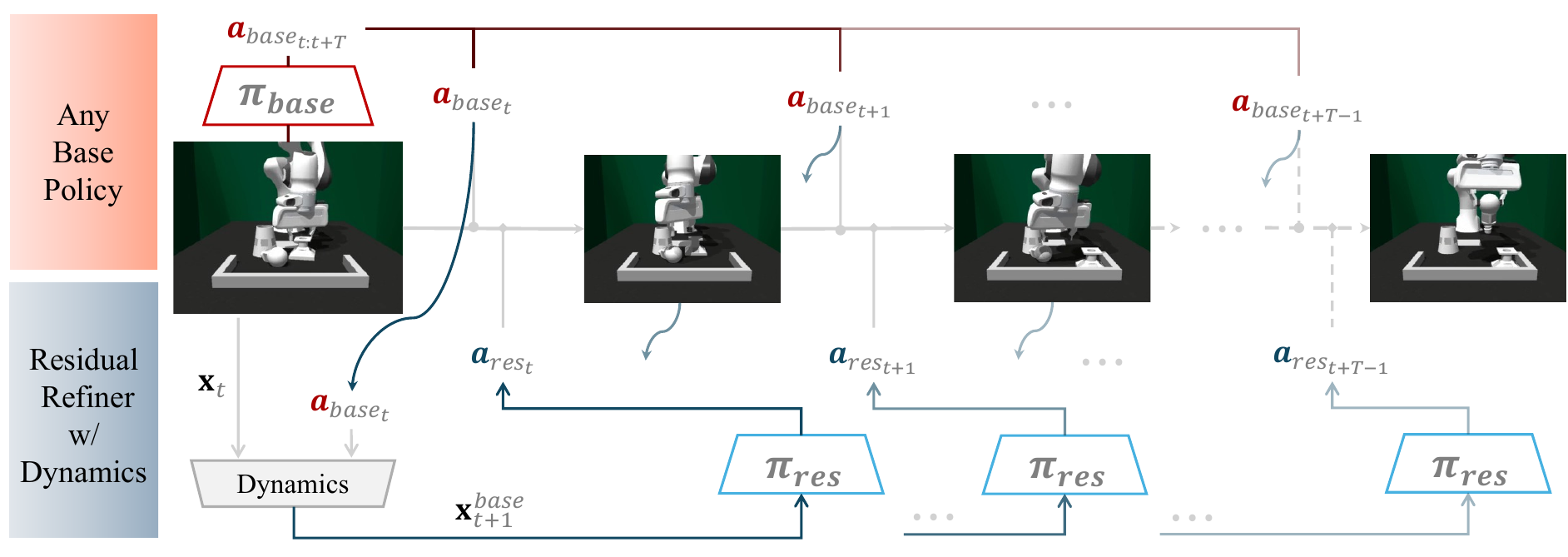}
    \caption{
    \textbf{Residual policy augmented with non-linear dynamics.} 
    An additional MLP-based dynamics model predicts the next state based on the base policy's action and the current state, providing the imagined state for the residual policy. The dynamics model is implemented as a two-layer MLP.
    }
    \label{fig:resip_w_nlin_dyn}
\end{figure}

To further explore the impact of dynamics modeling, we extend the standard residual policy~\citep{ankile2024imitation} by incorporating an additional non-linear dynamics model. Specifically, an MLP-based dynamics network is trained to predict the next state given the current state and the base policy's action, following a similar structure to KORR. 
The residual policy then conditions on the imagined next state generated by this dynamics model.
As illustrated in~\autoref{fig:resip_w_nlin_dyn}, the dynamics model is trained jointly during online reinforcement learning, with an auxiliary mean squared error (MSE) loss between the predicted and true next states. Note that the predicted states or observations maintain the same dimensionality as the original states or observations. The main paper presents detailed evaluations of this residual policy variant with non-linear dynamics in~\autoref{sec:experiment_linear_study}.

\section{Linear vs. Non-Linear}
\label{appendix:linear_vs_nonlinear}
We examine the choice between linear and non-linear dynamics for residual policy learning. While non-linear models offer higher expressiveness, they often lack stability and generalization under limited data. 
To assess this, we combine empirical comparisons with deeper non-linear architectures and theoretical analyses with formal proofs of their extrapolation behavior.

As reported in~\reftab{tab:linvsnonlin}, non-linear dynamics modeled by MLPs show no improvement and even cause a notable performance drop, despite using deeper, more expressive architectures. 
This is consistent with the findings in recent work on the benefits brought from linear modeling~\citep{mondal2023efficient, fujimoto2025towards}.
\begin{table*}[ht]
    \small
    \centering
    \begin{tabular}{l|cccc}
    \toprule
    Task & ResiP & Non-Linear (2-layer) & Non-Linear (4-layer) & KORR (Ours) \\
    \midrule
    \texttt{Low} & 98.14 & 97.32 & 97.47 & \textbf{98.73} ($\uparrow$) \\
    \texttt{Med} & 84.99 & 81.93 & 80.76 & \textbf{87.21} ($\uparrow$) \\
    \bottomrule
    \end{tabular}
    \caption{
    \textbf{Evaluation with stronger non-linear dynamics.} Results on the \texttt{One\_Leg} task without disturbance using Diffusion Policy~\citep{chi2023diffusion} as the base policy. Deeper non-linear models fail to improve and even reduce performance, while Koopman dynamics consistently improves results. 
    }
    \label{tab:linvsnonlin}
\end{table*}

\begin{figure}[ht]
    \centering
    \includegraphics[width=0.8\textwidth]{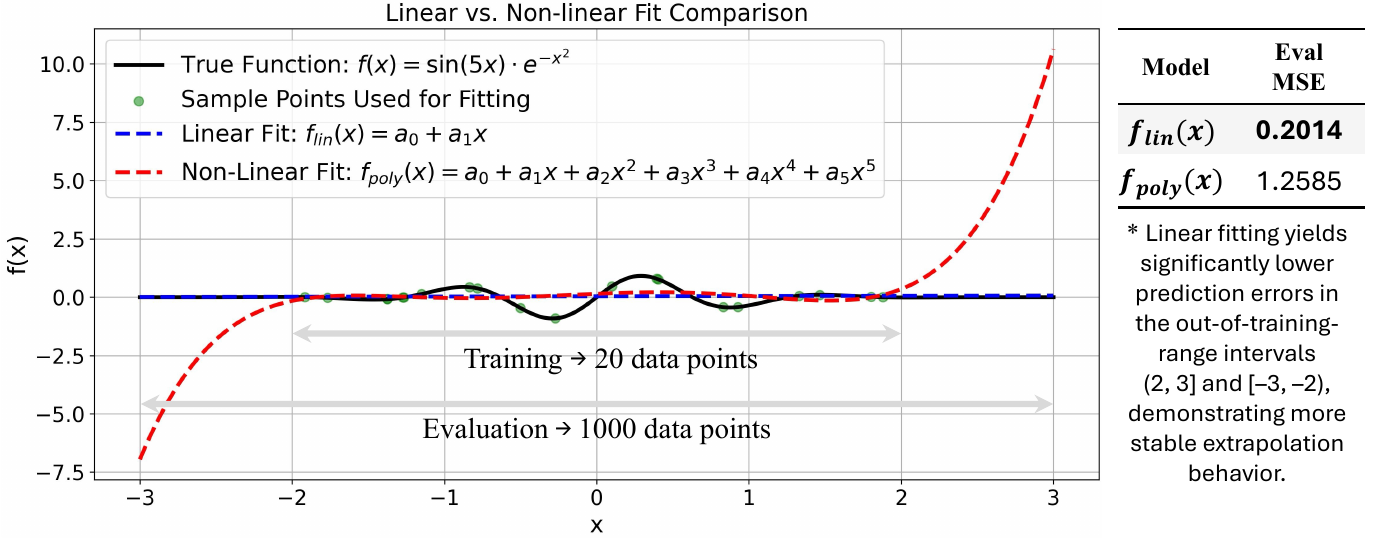}
    \caption{
        \textbf{Extrapolation behavior of linear vs. non-linear models.}
        A simple analytical experiment on polynomials demonstrates that linear models extrapolate conservatively and stably, while non-linear models suffer from large errors outside the training range. 
    }
    \label{fig:linvsnonlin}
\end{figure}

One explanation lies in extrapolation. 
When fitting complex non-linear targets with limited and range-restricted data, linear models often generalize better due to their conservative extrapolation behavior~\citep{fujimoto2025towards}. This advantage becomes more pronounced in long-horizon settings where errors accumulate over time.
As shown in~\autoref{fig:linvsnonlin}, we fit both linear and non-linear models to a complex polynomial target function using limited training data. 
Outside the training range (e.g., $(2,3]$), the non-linear model exhibits large prediction errors that worsen with increased extrapolation distance. 
In contrast, the linear model remains stable and achieves lower overall MSE, reflecting its robustness and superior generalization due to the conservative nature of linear extrapolation.

To further support the empirical results, we compare the extrapolation behavior of a linear model $\hat{f}_{\mathrm{lin}}(x) = a_0 + a_1 x$ and a non-linear polynomial $\hat{f}_{\mathrm{poly}}(x) = \sum_{i=0}^5 a_i x^i$ over $x \in [x_0, x_0 + \delta]$, where $x_0$ is at the edge of the training region (with limited dataset). The linear model has a constant derivative and is globally Lipschitz:
\begin{equation*}
    |\hat{f}_{\mathrm{lin}}(x) - \hat{f}{_\mathrm{lin}}(x_0)| = |a_1| \cdot |x - x_0|
\end{equation*}
The extrapolation grows linearly, predictably bounded by $|x - x_0|$: 
\begin{equation*}
|f(x) - \hat{f}_{\mathrm{lin}}(x)| \leq |f(x) - f(x_0)| + |f(x_0) - \hat{f}_{\mathrm{lin}}(x_0)| + |a_1| \cdot |x - x_0|
\end{equation*}
Taylor’s theorem bounds the error of the non-linear polynomial as:
\begin{equation*}
|f(x) - \hat{f}_{\mathrm{poly}}(x)| \leq \frac{|f^{(d+1)}(\xi)|}{(d+1)!} |x - x_0|^{d+1} \quad \text{for some } \xi \in [x_0, x]
\end{equation*}
The extrapolation error scales as $\mathcal{O}(|x - x_0|^{d+1})$, leading to rapid error amplification even with moderate $d$ and small extrapolation beyond the training region.
This highlights the instability of high-degree polynomials in extrapolation, whereas linear models provide more stable and conservative predictions, mitigating error amplification in long-horizon prediction settings.

\section{Additional Ablation Studies}
\label{appendix:add_abl_exp}
\begin{figure}[ht]
    \centering
    \includegraphics[width=0.8\textwidth]{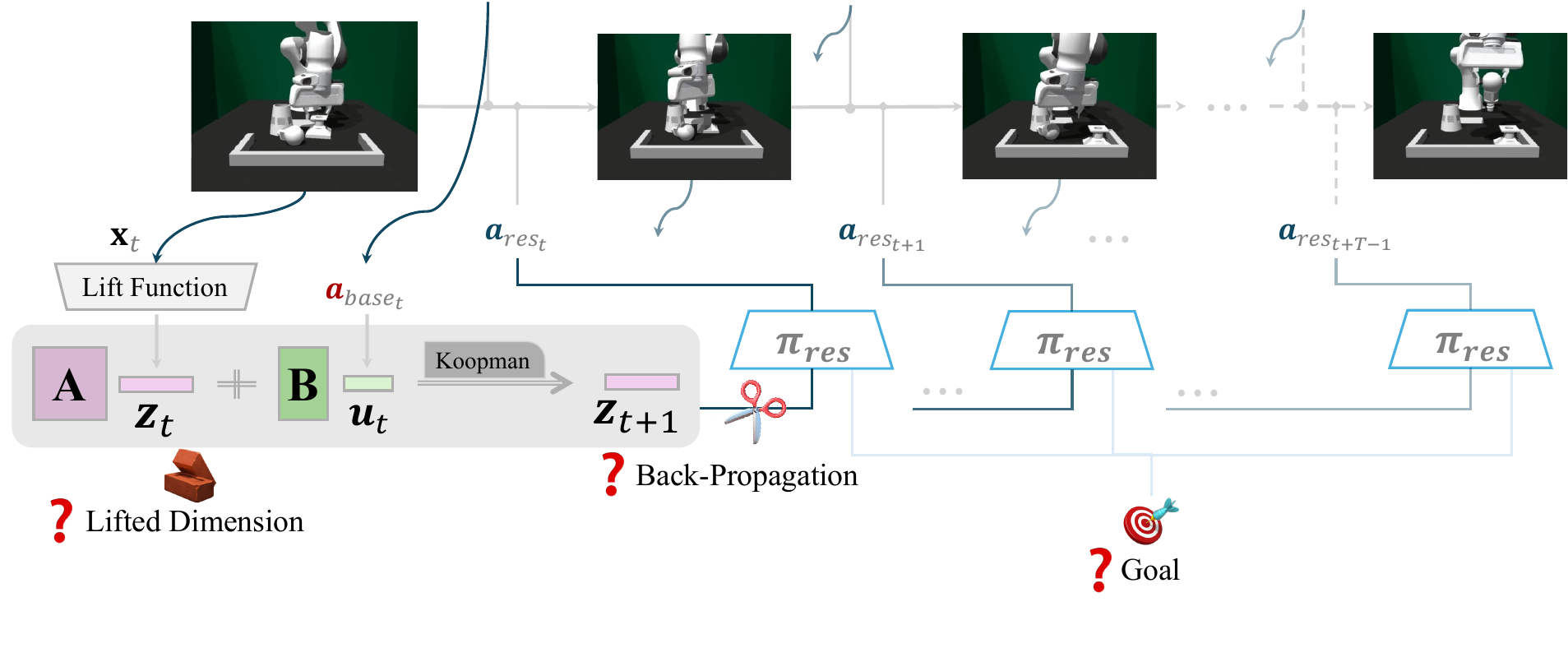}
    \vspace{-0.15in}
    \caption{
    \textbf{Visualization of key ablation studies in KORR.}
    Highlighted ‘question’ emojis indicate the specific ablation topics discussed below, such as lifted dimension and others.
    }
    \label{fig:korr_abl_vis}
\end{figure}

We conduct several ablation studies to further investigate the design choices of KORR. As illustrated in~\autoref{fig:korr_abl_vis}, we focus on three key aspects: the lifted dimension, the role of reinforcement learning (RL) loss in training the Koopman model, and the potential of integrating goal-conditioning into the residual policy, all of which highlight the contribution of Koopman modeling.

\subsection{Ablation on Lifted Dimension}
\label{appendix:add_abl_exp_liftdim}
\begin{figure}[ht]
  \begin{center}
    \includegraphics[width=0.52\linewidth]{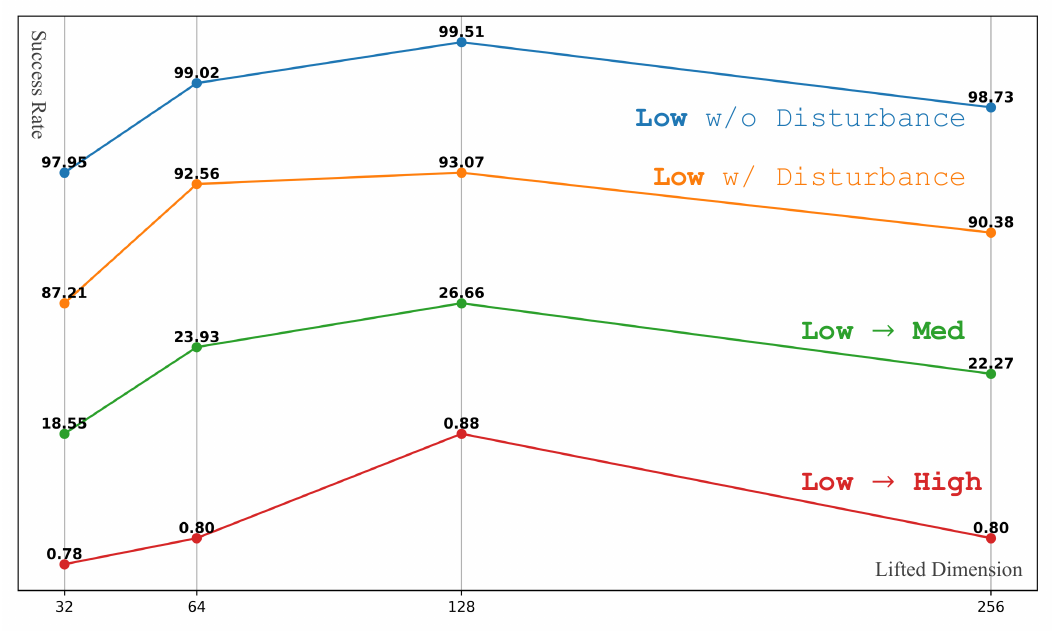}
  \end{center}
  \vspace{-0.1in}
  \caption{\textbf{Effect of lifted dimension.} Ablation study on \texttt{One\_Leg} under \texttt{Low} randomization.}
  \label{fig:abl_liftdim}
\end{figure}
We study the impact of the lifted dimension in the Koopman embedding. The lift function $g_{\theta}$ maps observations into a higher-dimensional latent space to enable linear dynamics modeling. While traditional approaches require manual design, we learn $g_{\theta}$ via a neural network.
As shown in~\autoref{fig:abl_liftdim}, we evaluate various lifted dimensions on the \texttt{One\_Leg} task, considering both initial randomization and evaluation disturbances. The original state consists of 58 dimensions, comprising 6 objects with 7-dimensional position representations each, along with 16-dimensional proprioceptive information.
A 32D lifted space underperforms across all metrics, indicating insufficient capacity. Larger dimensions (64D, 128D, 256D) perform similarly, suggesting that once essential information is captured, further expansion brings marginal benefit. Interestingly, the 128-dimensional lifted space slightly outperforms others, while excessively large dimensions introduce unnecessary complexity. This underscores the importance of selecting an appropriate lifted dimension tailored to task complexity.

\subsection{Ablation on RL Loss Backpropagation to Koopman Dynamics}
\label{appendix:add_abl_exp_bkp}
\begin{table*}[ht]
    \small
    \centering
    \setlength{\tabcolsep}{6pt}    
    \begin{tabular}{lcccccccc}
    \toprule
    \multirow{3}{*}{\textbf{Methods}} 
    & \multicolumn{6}{c}{\texttt{One\_Leg}} 
    & \multicolumn{2}{c}{\texttt{Round\_Table}} \\
    \cmidrule(r){2-7} 
    \cmidrule(r){8-9} 
    & \multicolumn{2}{c}{\texttt{Low}} 
    & \multicolumn{2}{c}{\texttt{Med}}  
    & \multicolumn{2}{c}{\texttt{High}} 
    & \multicolumn{2}{c}{\texttt{Low}} \\
    \cmidrule(r){2-3}
    \cmidrule(r){4-5}
    \cmidrule(r){6-7}
    \cmidrule(r){8-9}
    & \texttt{w/o} & \texttt{w/} & \texttt{w/o} & \texttt{w/} & \texttt{w/o} & \texttt{w/} & \texttt{w/o} & \texttt{w/} \\
    \midrule
    KORR & \textbf{98.73} & \textbf{90.38} & \textbf{87.21} & \textbf{52.31} & \textbf{44.04} & \textbf{8.30} & \textbf{96.78} & \textbf{81.35} \\
    KORR \texttt{w/o} bkp ($\downarrow$) & 95.99 & 83.59 & 73.73 & 35.16 & 19.24 & 2.83 & 64.94 & 28.91 \\
    \bottomrule
    \end{tabular}
    \caption{\textbf{Effect of disabling RL loss backpropagation to Koopman dynamics.} Removing RL feedback severely degrades performance across tasks and randomization levels.}
    \label{tab:abl_wo_bkp}
\end{table*}
In KORR, RL losses are allowed to backpropagate into the Koopman dynamics module alongside a standalone Koopman prediction loss (see~\refalg{algo:koopman_residual_policy}).
We ablate this design choice by disabling RL loss gradients to the Koopman model, conducting experiments on the \texttt{One\_Leg} and \texttt{Round\_Table} tasks across multiple randomization levels, with and without disturbance during evaluation.
As shown in~\reftab{tab:abl_wo_bkp}, performance consistently deteriorates across all settings.
This suggests that minimizing the Koopman prediction loss alone is insufficient; incorporating task rewards is essential to guide the lifted representation in retaining reward-relevant features for accurate linear predictions. 
Thus, RL-informed training is essential for learning meaningful, task-aligned dynamics in KORR, leading to enhanced performance and robustness. And we retain backpropagation of RL loss during training.

\subsection{Ablation on Goal-Conditioned Residual Policies}
\label{appendix:add_abl_exp_goal}
\begin{table*}[ht]
    \small
    \centering
    \setlength{\tabcolsep}{6pt}
    \begin{tabular}{lcccccccc}
    \toprule
    \multirow{3}{*}{\textbf{Methods}} 
    & \multicolumn{4}{c}{\texttt{One\_Leg}} 
    & \multicolumn{4}{c}{\texttt{Lamp}} \\
    \cmidrule(r){2-5} 
    \cmidrule(r){6-9} 
    & \multicolumn{2}{c}{\texttt{Low}} 
    & \multicolumn{2}{c}{\texttt{Med}}  
    & \multicolumn{2}{c}{\texttt{Low}} 
    & \multicolumn{2}{c}{\texttt{Med}} \\
    \cmidrule(r){2-3}
    \cmidrule(r){4-5}
    \cmidrule(r){6-7}
    \cmidrule(r){8-9}
    & \texttt{w/o} & \texttt{w/} & \texttt{w/o} & \texttt{w/} & \texttt{w/o} & \texttt{w/} & \texttt{w/o} & \texttt{w/} \\
    \midrule
    ResiP & \textbf{98.14} & \textbf{85.35} & \textbf{84.99} & \textbf{46.09} & \textbf{86.58} & \textbf{60.55} & \textbf{51.66} & \textbf{29.98} \\
    ResiP + Goal  ($\downarrow$)& 96.78 & 81.15 & 82.62 & 39.55  & 44.53 & 26.27 & 48.54 & 24.68\\
    \midrule
    KORR & 98.73 & \textbf{90.38} & 87.21 & 52.31 & 89.65 & 63.48 & 74.47 & 39.16 \\
    KORR + Goal ($\uparrow$) & \textbf{99.02} & 90.27 & \textbf{87.60} & \textbf{53.81} & \textbf{90.02} & \textbf{65.57} & \textbf{75.10} & \textbf{40.04} \\
    \bottomrule
    \end{tabular}
    \caption{\textbf{Effect of goal-conditioning on residual policies.} Goal information slightly improves KORR but harms ResiP, highlighting KORR's potential advantage in future goal-conditioned tasks.}
    \label{tab:abl_goal_cond}
\end{table*}
We investigate incorporating goal-conditioning into the residual policy by providing the final desired state as an additional input. Experiments are conducted on the \texttt{One\_Leg} and \texttt{Lamp} tasks under both \texttt{Low} and \texttt{Med} initial randomization levels, with evaluations performed both with and without disturbances to comprehensively assess the approach. The final goal state is constructed using the last frame of a successful rollout.
As shown in~\reftab{tab:abl_goal_cond}, adding goal information degrades the performance of ResiP but marginally improves KORR. In ResiP, goal-conditioning introduces a mismatch: the inputs (current state and base action) are not naturally aligned with goal states, increasing training difficulty. In contrast, KORR models the imagined next state under base actions, which shares the same space with the goal, making comparison natural. Thus, goal-conditioning helps KORR generate more goal-directed corrections, albeit the improvement is limited due to the single fixed goal setting. These results highlight the potential of KORR for flexible, goal-conditioned refinement.

\section{Influence of Base Policy Quality on Residual Learning}
\label{appendix:influ_base_to_residual}
\begin{wrapfigure}{r}{0.50\textwidth}
  \vspace{-4mm}
  \begin{center}
    \includegraphics[width=\linewidth]{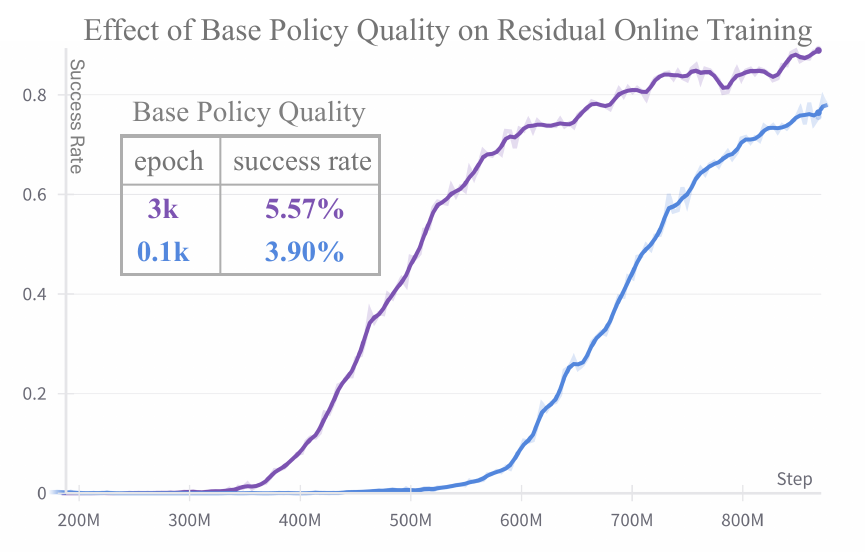}
  \end{center}
  \vspace{-2mm}
  \caption{\textbf{Effect of base policy quality on residual learning.}
  Residual policies trained on better-trained base policies learn faster and achieve higher final performance.
  }
  \label{fig:abl_base2res}
  \vspace{-4mm}
\end{wrapfigure}
We investigate how the quality of the base policy affects residual policy training and final task performance. As shown in~\autoref{fig:abl_base2res}, we compare the training curves of ResiP~\citep{ankile2024imitation} (which exhibits similar behavior within our KORR framework) on the \texttt{Lamp} task with \texttt{Low} initialization randomization, using the same base diffusion policy architecture but trained for different durations. Specifically, we consider two base policies: one trained for only 100 steps (undertrained) and one trained for 3,000 steps (well-trained).
The results show that a stronger base policy significantly accelerates the residual policy’s online learning, leading to faster convergence and higher final performance. Even when the base policy is imperfect, improvements in its quality provide a better foundation for residual exploration, reducing redundant search and increasing the likelihood of successful training.

\section{Impact of Exploration Rate on Residual Learning}
\label{appendix:explo_rate_res_learn}
\begin{figure}[ht]
  \begin{center}
    \includegraphics[width=0.96\linewidth]{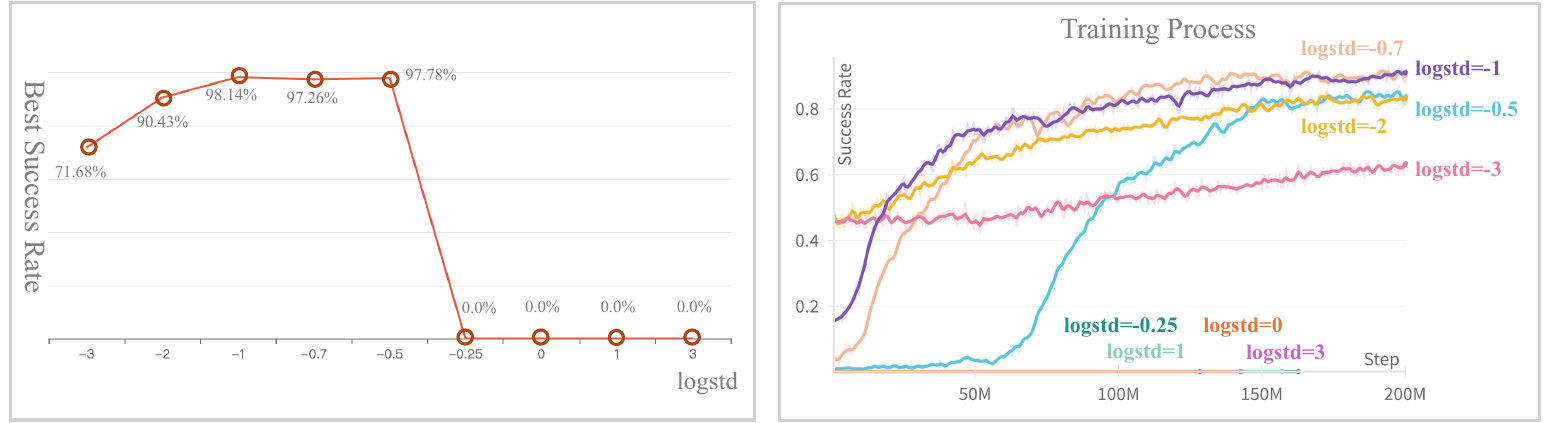}
  \end{center}
  \vspace{-0.1in}
  \caption{\textbf{Ablation on exploration rate.} Properly balancing exploration is crucial: excessive exploration destabilizes early learning, while insufficient exploration limits policy improvement. A moderate exploration rate enables stable and efficient residual policy training.}
  \label{fig:abl_logstd}
\end{figure}
We investigate the impact of exploration rate during the online reinforcement learning of the residual policy. 
Specifically, we control exploration by fixing the log standard deviation (\texttt{logstd}) of the action distribution throughout training.
As shown in~\autoref{fig:abl_logstd}, evaluated on the \texttt{One\_Leg} task under \texttt{Low} randomness, excessive exploration leads to unstable rollouts and early failures, preventing reward acquisition and hindering learning. 
Conversely, insufficient exploration limits the policy’s ability to discover better solutions, resulting in suboptimal performance.
Balancing these effects, we fix $\text{logstd} = -1$ across all experiments to ensure stable and effective training.

\section{Proximal Policy Optimization (PPO) Formulation}
\label{appendix:ppo_alg}

For online fine-tuning of the residual policy in our KORR framework, we employ Proximal Policy Optimization (PPO)~\citep{schulman2017proximal}, a widely adopted on-policy reinforcement learning algorithm known for its robustness and sample efficiency in continuous control tasks. The algorithm is used in the training loop outlined in~\refalg{algo:koopman_residual_policy}. PPO optimizes a composite objective composed of three terms: a clipped policy surrogate loss, a value function loss, and an entropy bonus. The following section details the complete formulation and rationale behind each term.

\textbf{Advantage estimation}, $A_t$, is computed using Generalized Advantage Estimation (GAE)~\citep{schulman2015high}, which balances bias and variance through the hyperparameter $\lambda$. It is calculated as:
\begin{equation}
    A_t = \sum_{l=0}^{T - t} (\gamma \lambda)^l \delta_{t+l}; \delta_t = r_t + \gamma V(\mathbf{x}_{t+1}) - V(\mathbf{x}_t),
\end{equation}
where $r_t$ is the scalar reward at time step $t$, $\gamma$ is the discount factor, and $V(\cdot)$ denotes the learned value function. This estimation is performed in line 12 of~\refalg{algo:koopman_residual_policy} after collecting the rollout buffer $\mathcal{D}$. Here, $\mathbf{x}_t$ denotes the full system state at time $t$, which may also correspond to raw observations.

\textbf{Clipped policy surrogate loss} ensures training stability and avoids large policy updates. PPO uses a clipped surrogate loss:
\begin{equation}
    \mathcal{L}^{\text{CLIP}}(\phi) = \mathbb{E}_t \left[ \min \left( r_t(\phi) A_t, \text{clip}(r_t(\phi), 1 - \epsilon, 1 + \epsilon) A_t \right) \right],
\end{equation}
with the importance sampling ratio defined as:
\begin{equation}
    r_t(\phi) = \frac{\pi_{\phi}(\boldsymbol{a}_t | \mathbf{x}_t)}{\pi_{\phi_{\text{old}}}(\boldsymbol{a}_t | \mathbf{x}_t)}.
\end{equation}
This formulation penalizes updates where the new policy deviates too far from the old one. The clipping parameter $\epsilon$ is a small scalar hyperparameter, and $\pi_{\phi}$ denotes the residual policy $\pi_{\text{res}}$ parameterized by $\phi$.

\textbf{Value function loss} enforces accurate prediction of expected returns, which is crucial for computing advantages:
\begin{equation}
    \mathcal{L}^{\text{VF}}(\phi) = \mathbb{E}_t \left[ \left( V_\phi(\mathbf{x}_t) - \hat{V}_t \right)^2 \right],
\end{equation}
where the target value $\hat{V}_t$ is computed as:
\begin{equation}
    \hat{V}_t = A_t + V(\mathbf{x}_t).
\end{equation}
This formulation allows the critic to learn a bootstrapped estimate of future rewards.

\textbf{Entropy bonus} encourages exploration and prevents premature convergence to deterministic policies. PPO includes an entropy regularization term:
\begin{equation}
    \mathcal{L}^{\text{S}}(\phi) = \mathbb{E}_t \left[ -\pi_\phi(\boldsymbol{a}_t | \mathbf{x}_t) \log \pi_\phi(\boldsymbol{a}_t | \mathbf{x}_t) \right].
\end{equation}
This is particularly important in high-dimensional and stochastic environments, such as robot manipulation tasks with contact dynamics and sensor noise.

\textbf{Final objective} is a weighted combination of the above components:
\begin{equation}
    \mathcal{L}_{\text{ppo}}(\phi) = \mathcal{L}^{\text{CLIP}}(\phi) - c_1 \mathcal{L}^{\text{VF}}(\phi) + c_2 \mathcal{L}^{\text{S}}(\phi),
\end{equation}
where $c_1$ and $c_2$ are scalar weights that balance the contributions of the value function loss and entropy regularization, respectively. This overall PPO objective $\mathcal{L}_{\text{ppo}}$ corresponds to line 15 in~\refalg{algo:koopman_residual_policy}, and is used to update all residual policy parameters via gradient descent.

\end{document}